\documentclass[journal]{IEEEtran}

\usepackage{amsfonts,amssymb,amsmath,amstext,bm}
\usepackage{epsf,graphicx}
\graphicspath{{./}}  

\usepackage{url}   
\usepackage{cite}  

\def\vec#1{\mbox{\boldmath$\displaystyle#1$}}
\def\e    {\displaystyle{e}}
\def\b    {\vec{b}}
\def\c    {\vec{c}}
\def\x    {\vec{x}}
\def\xx   {\stackrel{\thicksim}{\vec{x}}}
\def\y    {\vec{y}}
\def\h    {\vec{h}}
\def\W    {\vec{W}}

\title{A Neighbourhood-Based Stopping Criterion for Contrastive Divergence Learning}

\author{Enrique~Romero~Merino and Ferran~Mazzanti~Castrillejo and Jordi~Delgado~Pin
\thanks{Enrique~Romero~Merino is with the Departament de Ci\`encies de
  la Computaci\'o, Universitat Polit\`ecnica de Catalunya, Spain (e-mail:
  eromero@cs.upc.edu).}
\thanks{Ferran~Mazzanti~Castrillejo is with the Departament de
  F\'{\i}sica i Enginyeria Nuclear, Universitat Polit\`ecnica de
  Catalunya, Spain (e-mail: ferran.mazzanti@upc.edu).}
\thanks{Jordi~Delgado~Pin is with the Departament Ci\`encies de
  la Computaci\'o, Universitat Polit\`ecnica de Catalunya, Spain
  (e-mail: jdelgado@cs.upc.edu).}
}

\begin{document} 

\maketitle

\begin{abstract} 
  Restricted Boltzmann Machines (RBMs) are general unsupervised
  learning devices to ascertain generative models of data
  distributions. RBMs are often trained using the Contrastive
  Divergence learning algorithm (CD), an approximation to the gradient
  of the data log-likelihood. A simple reconstruction error is often
  used as a stopping criterion for CD, although several
  authors~\cite{schulz-et-al-Convergence-Contrastive-Divergence-2010-NIPSw,
    fischer-igel-Divergence-Contrastive-Divergence-2010-ICANN} have
  raised doubts concerning the feasibility of this procedure.  In many
  cases the evolution curve of the reconstruction error is monotonic
  while the log-likelihood is not, thus indicating that the former is
  not a good estimator of the optimal stopping point for learning.
  However, not many alternatives to the reconstruction error have been
  discussed in the literature. In this manuscript we investigate
  simple alternatives to the reconstruction error, based on the
  inclusion of information contained in neighboring states to the
  training set, as a stopping criterion for CD learning.
\end{abstract} 


\section{Introduction}
\label{introduction}

Learning algorithms for deep multilayer neural networks have been
known for a long time \cite{RUMELHART-HINTON-WILLIAMS86}, 
though they usually could not outperform simpler, shallow networks.
In this way, deep multilayer networks
were not widely used to solve large scale real-world
problems until the last decade~\cite{bengio-DeepArchitectures-2009-bk}.
In 2006, Deep Belief Networks (DBNs)
\cite{hinton-et-al-DeepBeliefNetworks-2006-NC} came out as a real
breakthrough in this field, since the learning algorithms proposed
ended up being a feasible and practical method to train deep
networks, with spectacular results
\cite{hinton-salakhutdinov-Auto-Encoders-2006-Science,
larochelle-et-al-Strategies-Train-Deep-Networks-2009-JMLR,
lee-at-al-Convolutional-Scalable-2009-ICML,le-at-al-google_unsupervised-2012-ICML}. DBNs
have Restricted Boltzmann Machines (RBMs)
\cite{smolensky-Restricted-Boltzmann-Machines-1986-PDP} as their
building blocks.

RBMs are topologically constrained Boltzmann Machines
(BMs) with two layers, one of hidden and another of visible neurons, and no
intralayer connections. This property makes working with RBMs simpler than with regular
BMs, and in particular the stochastic computation of the log-likelihood gradient may be
performed more efficiently by means of Gibbs sampling
\cite{geman-geman-Gibbs-Sampling-1984-TPAMI,bengio-DeepArchitectures-2009-bk}.

In 2002, the \textit{Contrastive Divergence} (CD) learning algorithm 
was proposed as an efficient training method for product-of-expert
models, from which RBMs are a special case
\cite{hinton-Contrastive-Divergence-2002-NC}. It was observed that
using CD to train RBMs worked quite well in practice. This fact was
important for deep learning since some authors suggested that a
multilayer deep neural network is better trained when each layer is
pre-trained separately as if it were a single RBM
\cite{hinton-salakhutdinov-Auto-Encoders-2006-Science,
bengio-et-al-ContinuousInputs-And-AutoEncoders-2007-NIPS,
larochelle-et-al-Strategies-Train-Deep-Networks-2009-JMLR}. Thus,
training RBMs with CD and stacking up them seems to be a good way to
go when designing deep learning architectures.

However, the picture is not as nice as it looks, since CD is not a
flawless training algorithm. Despite CD being an approximation of the
true log-likelihood gradient
\cite{bengio-delalleau-Justifying-ContrastiveDivergence-2009-NC}, it
is biased and it may not converge in some cases
\cite{carreira-hinton-Contrastive-Divergence-Learning-2005-AISTATS,
  yuille-Convergence-Contrastive-Divergence-2005-NIPS,
  mackay-Failures-Contrastive-Divergence-2001-TR}. Moreover, it has
been observed that CD, and variants such as Persistent CD
\cite{tieleman-Persistent-Contrastive-Divergence-2008-ICML} or Fast
Persistent CD
\cite{tieleman-hinton-Improved-Persistent-Contrastive-Divergence-2009-ICML}
can lead to a steady decrease of the log-likelihood during learning
\cite{fischer-igel-Divergence-Contrastive-Divergence-2010-ICANN,
  desjardins-et-al-Parallel-Tempering-2010-AISTATS}. Therefore, the
risk of learning divergence imposes the requirement of a stopping
criterion.  There are two main methods used to decide when to stop the
learning process. One is based on the monitorization of the
\textit{reconstruction
  error}~\cite{hinton-Practical-Guide-RBMs-2012-TT}. The other is
based on the estimation of the log-likelihood with \textit{Annealed
  Importance Sampling}
(AIS)~\cite{neal-Annealed-Importance-Sampling-1998-TR,
  salakhutdinov-et-al-Annealed-Importance-Sampling-2008-ICML}.  The
reconstruction error is easy to compute and it has been often used in
practice, though its adequacy remains unclear because of
monotonicity~\cite{fischer-igel-Divergence-Contrastive-Divergence-2010-ICANN}.
AIS seems to work better than the reconstruction error in most cases,
though it is considerably more expensive to compute, and may also fail
\cite{schulz-et-al-Convergence-Contrastive-Divergence-2010-NIPSw}.

In this work we approach this problem from a completely different
perspective. Based on the fact that the energy is a continuous and
smooth function of its variables, the close neighborhood of the
high-probability states is expected to acquire also a significant
amount of probability. In this sense, we argue that the information
contained in the neighborhood of the training data is valuable, and
that it can be incorporated in the learning process of RBMs.  In
particular, we propose to use it in the monitorization of the
log-likelihood of the model by means of a new quantity that depends on
the information contained in the training set and its neighbors.
Furthermore, and in order to make it computationally tractable, we
build it in such a way that it becomes independent of the partition
function of the model.  In this way, we propose a neighborhood-based
stopping criterion for CD and show its performance in several data
sets.  

\section{Learning in Restricted Boltzmann Machines}
\label{learning}

\subsection{Energy-based Probabilistic Models}

Energy-based probabilistic models define a probability distribution
from an energy function, as follows:
\begin{equation}
\label{pdf-energy-xh}
 P(\x,\h) = \frac{\e^{-\text{Energy}(\x,\h)}}{Z} \ ,
\end{equation}
where $\x$ and $\h$ stand for (typically binary) visible and hidden
variables, respectively.  The normalization factor $Z$ is called
partition function and reads
\begin{equation}
 Z = \sum_{\x,\h} \e^{-\text{Energy}(\x,\h)} \ .
\end{equation}

Since only $\x$ is observed, one is interested in the
marginal distribution
\begin{equation}
\label{pdf-energy-x-sumh}
 P(\x) = \frac{\sum_{\h} \e^{-\text{Energy}(\x,\h)}}{Z} \ ,
\end{equation}
but the evaluation of the partition function $Z$ is
computationally prohibitive since it involves an exponentially large number
of terms. In this way, one can not measure directly $P(\x)$.

The energy function depends on several parameters $\theta$, that are adjusted
at the learning stage. This is done by maximizing the
likelihood of the data. In energy-based models, the derivative of the
log-likelihood can be expressed as
\begin{eqnarray}
\label{dlog-likelihood}
\lefteqn{-\frac{\partial\log P(\x;\theta)}{\partial\theta} =
  \ E_{P(\h|\x)} \left[\frac{
      \partial\text{Energy}(\x,\h)}{\partial\theta}\right]} \nonumber
\\ {} & \ \ \ \ \ \ \ \ -\ E_{P(\xx)}
\left[E_{P(\h|\xx)}\left[\frac{\partial\text{Energy}(\xx,\h)}{\partial\theta}\right]
  \right] \ ,
\end{eqnarray}
where the first term is called the positive phase and the second term
the negative phase.  Similar to (\ref{pdf-energy-x-sumh}),
the exact computation of the derivative of the log-likelihood is
usually unfeasible because of the negative phase in
(\ref{dlog-likelihood}), which comes from the derivative of the
partition function.

\subsection{Restricted Boltzmann Machines}

Restricted Boltzmann Machines are
energy-based probabilistic models whose energy function is:
\begin{equation}
\label{energy-RBM-discrete-binary-binary}
 \text{Energy}(\x,\h) = -\b^t\x - \c^t\h - \h^t\W\x \ .
\end{equation}
RBMs are at the core of DBNs
\cite{hinton-et-al-DeepBeliefNetworks-2006-NC} and other deep
architectures that use RBMs for unsupervised pre-training previous to
the supervised step
\cite{hinton-salakhutdinov-Auto-Encoders-2006-Science,
  bengio-et-al-ContinuousInputs-And-AutoEncoders-2007-NIPS,
  larochelle-et-al-Strategies-Train-Deep-Networks-2009-JMLR}.

The consequence of the particular form of the energy function is that
in RBMs both $P(\h|\x)$ and $P(\x|\h)$ factorize. In this way it is
possible to compute $P(\h|\x)$ and $P(\x|\h)$ in one step, making it
possible to perform Gibbs sampling efficiently,
in contrast to  more general models like Boltzmann
Machines~\cite{aarts-korst-BoltzmannMachines-1990-book}.

\subsection{Contrastive Divergence}
\label{cd}
The most common learning algorithm for RBMs uses an algorithm to
estimate the derivative of the log-likelihood of a Product of Experts
model. This algorithm is called
Contrastive Divergence~\cite{hinton-Contrastive-Divergence-2002-NC}.

Contrastive Divergence CD$_n$ estimates the derivative of the
log-likelihood for a given point $\x$ as
\begin{eqnarray}
\label{dlog-likelihood-CD-n}
\lefteqn{-\frac{\partial\log P(\x;\theta)}{\partial\theta} \simeq
  \ E_{P(\h|\x)} \left[\frac{
      \partial\text{Energy}(\x,\h)}{\partial\theta}\right]}
\nonumber \\ {} &
\ \ \ \ \ \ \ \ -\ E_{P(\h|\x_{n})}\left[\frac{\partial\text{Energy}(\x_{n},\h)}{\partial\theta}\right]
\ .
\end{eqnarray}
where $\x_{n}$ is the last sample from the Gibbs chain starting from
$\x$ obtained after $n$ steps:
\begin{itemize}
\item[] $\h_1 \sim P(\h|\x)$
\item[] $\x_1 \sim P(\x|\h_1)$
\item[] ...
\item[] $\h_n \sim P(\h|\x_{n-1})$
\item[] $\x_{n} \sim P(\x|\h_n)$ \ .
\end{itemize}

Usually, 
$E_{P(\h|\x)}\left[\frac{\partial\text{Energy}(\x,\h)}{\partial\theta}\right]$
can be easily computed.

Several alternatives to CD$_n$ are Persistent CD 
\cite{tieleman-Persistent-Contrastive-Divergence-2008-ICML}, Fast
Persistent CD
\cite{tieleman-hinton-Improved-Persistent-Contrastive-Divergence-2009-ICML}
or Parallel Tempering
\cite{desjardins-et-al-Parallel-Tempering-2010-AISTATS}.

\subsection{Monitoring the Learning Process in RBMs}

Learning in RBMs is a delicate procedure involving a lot of data
processing that one seeks to perform at a reasonable speed in
order to be able to handle large spaces with a huge amount of
states. In doing so, drastic approximations that can only be
understood in a statistically averaged sense are 
performed~\cite{fischer-igel-RestrictedBoltzmannMachines-2014-PR}.

One of the most relevant points to consider at the learning stage is
to find a good way to determine whether a good solution has been found
or not, and so to decide when the learning process should stop. One
of the most widely used criteria for stopping is based on the
monitorization of the reconstruction error, which is a measure of the
capability of the network to produce an output that is consistent with
the data at input. Since RBMs are probabilistic models, the
reconstruction error of a data point $\x^{(i)}$ is computed as the
probability of $\x^{(i)}$ given the expected value of $\h$ for
$\x^{(i)}$:
\begin{equation}
\label{reconstruction-error-RBM-probability}
 R(\x^{(i)}) = -\log P\left(\x^{(i)} | E\left[\h|\x^{(i)}\right]\right) \ ,
\end{equation}
which is a probabilistic extension of the sum-of-squares reconstruction
error for deterministic networks
\begin{equation}
\label{rec-error-squares}
\epsilon(\x^{(i)}) = || \x^{(i)} - \x_{n}^{(i)} ||^2 \ .
\end{equation}

Some authors have shown that, in some cases, learning induces an
undesirable decrease in likelihood that goes undetected by the
reconstruction error
\cite{schulz-et-al-Convergence-Contrastive-Divergence-2010-NIPSw,
  fischer-igel-Divergence-Contrastive-Divergence-2010-ICANN}. It has
been shown
\cite{fischer-igel-Divergence-Contrastive-Divergence-2010-ICANN} that
the reconstruction error defined in
(\ref{reconstruction-error-RBM-probability}) usually decreases
monotonically. Since no increase in the reconstruction error takes
place during training there is no apparent way to detect the change of
behavior of the log-likelihood for CD$_n$.

\section{Proposed Stopping Criterion}
\label{proposed-stopping-criterion}

The proposed stopping criterion is based on the monitorization of the
ratio of two quantities: the geometric average of the probabilities of
the training set, and the sum of probabilities of points in a given
neighbourhood of the training set. More formally, what we monitor is
\begin{equation}
\label{PXY_ek_prob_b1}
\xi_d = \frac{\left[\prod_{i=1}^NP(\x^{(i)})\right]^{1/N}}
{{1\over |D|}\sum_{j\in D}P(\y^{(j)})} \ ,
\end{equation}
where $D$ is a subset of points at a Hamming distance from the
training set less or equal than $d$.
The idea behind the definition is that the evolution of $\xi_d$ at the
learning stage is expected to closely resemble that of the
log-likelihood for certain values of $d$ and $D$. For that reason we
propose as the stopping criterion to find the maximum of $\xi_d$,
which will be close to the one shown by the log-likelihood of the data,
as shown by the experiments in the next sections.

The reason for that is twofold. On one hand the numerator and
denominator monitor different things. The numerator, which is
essentially the likelihood of the data, is sensitive to the
accumulation of most of the probability mass by a reduced subset of
the training data, a typical feature of CD$_n$. For continuity
reasons, the denominator is strongly correlated with the sum of
probabilities of the training data. Once the problem has been learnt,
the probabilities in a close neighborhood of the training set will be
high. The value of $\xi_d$ results from a delicate equilibrium between
these two quantities (see section~\ref{experiments}), which we
propose to use as a stopping criterion for learning.  On the other
hand,
due to the structure of $\xi_d$, the partition functions $Z$ involved
in both the numerator and denominator cancels out, which is a
necessary condition in the design of the quantity being monitorized. In other
words, the computation of $\xi_d$ can be equivalently defined as
\begin{equation}
\label{PXY_ek_prob_b1_efficient}
\xi_d = 
\frac{ \left[ \prod_{i=1}^N \sum_{\h}\e^{-\text{Energy}(\x^{(i)},\h)} \right]^{1/N}}
       {{1\over|D|}\sum_{j\in D}\sum_{\h}\e^{-\text{Energy}(\y^{(j)},\h)}} \ .
\end{equation}
The particular topology of RBMs allows to compute
$\sum_{\h}\e^{-\text{Energy}(\x,\h)}$ efficiently. This fact
dramatically decreases the computational cost involved in the
calculation, which would otherwise become unfeasible in most
real-world problems where RBMs could been successfully applied.

While the numerator in $\xi_d$ is directly evaluated from the data in
the training set, the problem of finding suitable values for $\y$
still remains. Indeed, the set of points at a given Hamming
distance $d$ from the training set
is independent of the weights and bias of
the network. In this way, it can be built once at the very beginning of the
process and be used as required during learning.
Therefore, two issues have to be sorted out before the criterion can
be applied. The first one is to decide a suitable value of $d$.
Experiments with different problems show that this is indeed problem
dependent, as is illustrated in the experimental section
below. The second one is the choice of the subset $D$, which strongly
depends on the size of the space being explored. 
For small spaces one can safely use the complete set of points at
a distance less than or equal to $d$, but that can be forbiddingly large in real world
problems. For this reason we explore two possibilities: one including
all points and another including only a random subset of the same size
as the training set, which is only as expensive as dealing with the
training set.

\section{Experiments}
\label{experiments}

We performed several experiments to explore the aforementioned
criterion defined in section~\ref{proposed-stopping-criterion} and
study the behavior of $\xi_d$ in comparison with the log-likelihood
and the reconstruction error of the data in several problems.  We have
explored problems of a size such that the log-likelihood can be
exactly evaluated and compared with the proposed $\xi_d$ parameter.

The first problem, denoted {\em Bars and Stripes} (BS), tries to
identify vertical and horizontal lines in 4$\times$4 pixel images. The
training set consists in the whole set of images containing all
possible horizontal or vertical lines (but not both), ranging from no
lines (blank image) to completely filled images (black image), thus
producing $2\times 2^4-2=30$ different images (avoiding the repetition
of fully back and fully white images) out of the space of $2^{16}$
possible images with black or white pixels.  The second problem, named
{\em Labeled Shifter Ensemble} (LSE), consists in learning 19-bit
states formed as follows: given an initial 8-bit pattern, generate
three new states concatenating to it the bit sequences 001, 010 or
100. The final 8-bit pattern of the state is the original one shifting
one bit to the left if the intermediate code is 001, copying it
unchanged if the code is 010, or shifting it one bit to the right if
the code is 100. One thus generates the training set using all
possible $2^8\times 3 = 768$ states that can be created in this form,
while the system space consists of all possible $2^{19}$ different
states one can build with 19 bits.  These two problems have already
been explored in
\cite{fischer-igel-Divergence-Contrastive-Divergence-2010-ICANN} and
are adequate in the current context since, while still large, the
dimensionality of space allows for a direct monitorization of the
partition function and the log-likelihood during learning.  For the
sake of completeness, we have also tested the proposed criterion on
randomly generated problems with different space dimensions, where
the number of states to be learnt is significantly smaller than the
size of the space. In particular, we have generated four different
data sets (RAN10, RAN12, RAN14 and RAN16) 
consisting of $N_v=10, 12, 14, 16$ binary input units and
$2^{N_v/2}$ examples to be learnt, as suggested
in~\cite{buhlmann-de-geer-Statistics-High-Dimensional-2011-book}.

In the following we discuss the learning processes of these problems
with binary RBMs, employing the Contrastive Divergence algorithm
CD$_n$ with $n=1$ and $n=10$ as described in section~\ref{cd}.  In the
BS case the RBM had 16 visible and 8 hidden units, while in the LSE
problem these numbers were 19 and 10, respectively.  For the random
data sets we have used 10 hidden units in each case.

Every simulation was carried out for a total of 50000 epochs, with
measures being taken every 50 epochs. Moreover, every point in the
subsequent plots was the average of ten different simulations starting
from different random values of the weights and bias. Other parameters
affecting the results that were changed along the analysis are the
learning rates involved in the weight and bias update rules.
No weight decay was used, and momentum was set to 0.8. 
The learning rates were chosen in order to make sure that the
log-likelihood degenerates, in such a way that it presents a clear
maximum that should be detected by $\xi_d$.

In the following we perform two series of experiments that are
reported in the next two subsections. In the first one
(section~\ref{complete}) we analyze the case where all states in $D$
are included.  In the second one~(section~\ref{uncomplete}) we relax
the computational cost of the evaluation of $\xi_d$ by selecting only
a small subset of all the states in $D$.

\subsection{Complete Neighborhoods}
\label{complete}

\begin{figure*}[t!]
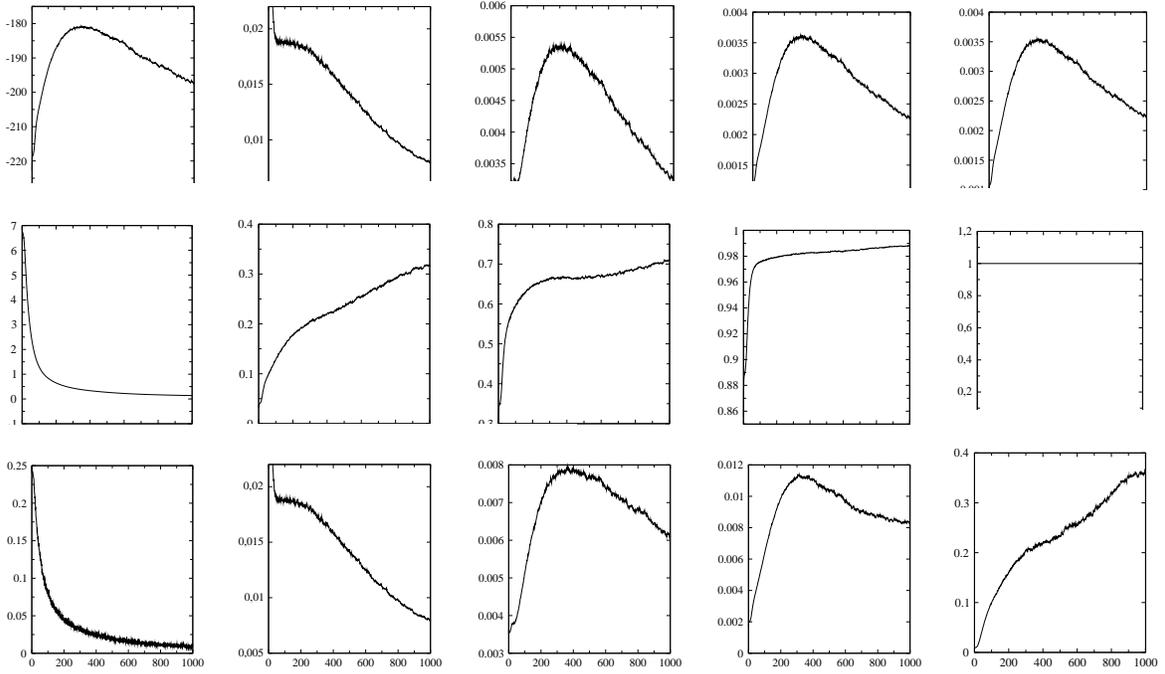

\begin{center}
\begin{tabular}{ccccc}
\includegraphics[width=0.147\textwidth] {RAN10_loglikelihood_Fig1_11.eps} &
\includegraphics[width=0.15\textwidth] {RAN10_xi_DA_d0_Fig1_12.eps} &
\includegraphics[width=0.155\textwidth] {RAN10_xi_DA_d1_Fig1_13.eps} &
\includegraphics[width=0.15\textwidth] {RAN10_xi_DA_d2_Fig1_14.eps} &
\includegraphics[width=0.15\textwidth] {RAN10_xi_DA_d3_Fig1_15.eps} \\
\includegraphics[width=0.145\textwidth] {RAN10_ProbRecErr_Fig1_21.eps} &
\includegraphics[width=0.15\textwidth] {RAN10_SumProbs_d0_Fig1_22.eps} &
\includegraphics[width=0.15\textwidth] {RAN10_SumProbs_d1_Fig1_23.eps} &
\includegraphics[width=0.15\textwidth] {RAN10_SumProbs_d2_Fig1_24.eps} &
\includegraphics[width=0.145\textwidth] {RAN10_SumProbs_d3_Fig1_25.eps} \\
\includegraphics[width=0.145\textwidth] {RAN10_SqrRecErr_Fig1_31.eps} &
\includegraphics[width=0.15\textwidth] {RAN10_xi_DS_d0_Fig1_32.eps} &
\includegraphics[width=0.15\textwidth] {RAN10_xi_DS_d1_Fig1_33.eps} &
\includegraphics[width=0.15\textwidth] {RAN10_xi_DS_d2_Fig1_34.eps} &
\includegraphics[width=0.15\textwidth] {RAN10_xi_DS_d3_Fig1_35.eps}
\end{tabular}
\end{center}
\caption{Results for the RAN10 problem. The first column shows the
  log-likelihood (top) and the reconstruction
  errors~(\ref{reconstruction-error-RBM-probability})
  and~(\ref{rec-error-squares}) (center and bottom). The other columns
  in the first, second and third rows depict $\xi_d$ for $D=D_A$, the
  sum of probabilities in the denominator of $\xi_d$ for $D=D_A$, and $\xi_d$ for
  $D=D_S$ for $d=0,1,2,3$, respectively. The x-axis is the number of
  epochs along the simulation divided by 50 in all plots. All data in
  the y-axis are in arbitrary units.}
\label{fig_RAN10}
\end{figure*}

\begin{figure*}[t!]
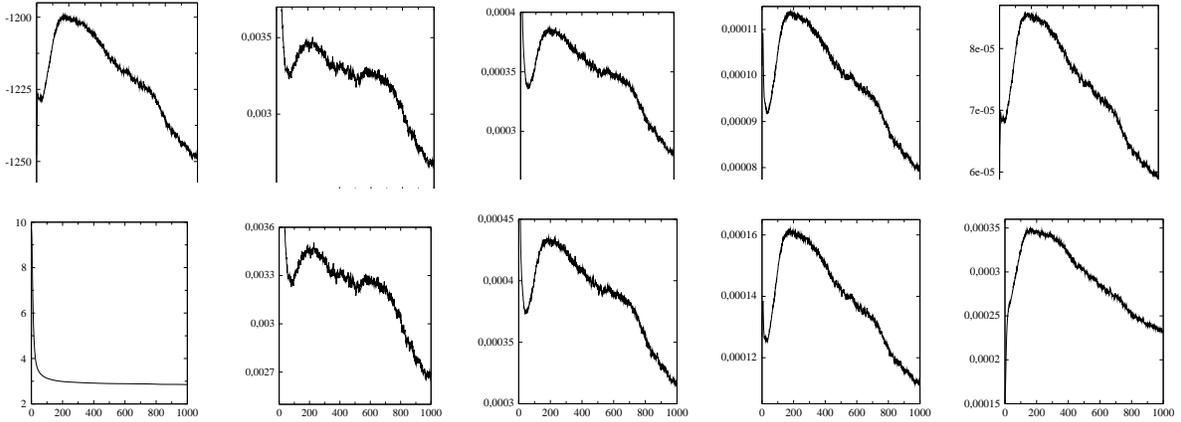

\begin{center}
\begin{tabular}{ccccc}
\includegraphics[width=0.15\textwidth] {RAN14_loglikelihood_Fig2_11.eps} &
\includegraphics[width=0.15\textwidth] {RAN14_xi_DA_d0_Fig2_12.eps} &
\includegraphics[width=0.15\textwidth] {RAN14_xi_DA_d1_Fig2_13.eps} &
\includegraphics[width=0.155\textwidth] {RAN14_xi_DA_d2_Fig2_14.eps} &
\includegraphics[width=0.1475\textwidth] {RAN14_xi_DA_d3_Fig2_15.eps} \\
\includegraphics[width=0.135\textwidth] {RAN14_ProbRecErr_Fig2_21.eps} &
\includegraphics[width=0.145\textwidth] {RAN14_xi_DS_d0_Fig2_22.eps} &
\includegraphics[width=0.155\textwidth] {RAN14_xi_DS_d1_Fig2_23.eps} &
\includegraphics[width=0.155\textwidth] {RAN14_xi_DS_d2_Fig2_24.eps} &
\includegraphics[width=0.155\textwidth] {RAN14_xi_DS_d3_Fig2_25.eps}
\end{tabular}
\end{center}
\caption{Results for the RAN14 problem. The first column shows the
  log-likelihood and the reconstruction
  error~(\ref{reconstruction-error-RBM-probability}) (top and bottom
  panels). The other columns in the upper and lower rows show $\xi_d$
  for $D=D_A$ and $\xi_d$ for $D=D_S$ for $d=0,1,2,3$, respectively.}
\label{fig_RAN14}
\end{figure*}

We present the results for the problems at hand, showing for each
analyzed instance different plots corresponding to the actual
log-likelihood of the problem and $\xi_d$ for different values of $d$,
among other things. In order to identify the contributions to $\xi_d$
from the different neighborhoods of the training set,
we define two different sets: $D_A$ 
containing all states at a distance less than or equal to $d$, and
$D_S$ accounting for those states at a distance exactly equal to $d$.
We have computed $\xi_d$ for $D=D_A$ and $D=D_S$ in all our
experiments that are commented in the following.

\begin{figure*}[t!]
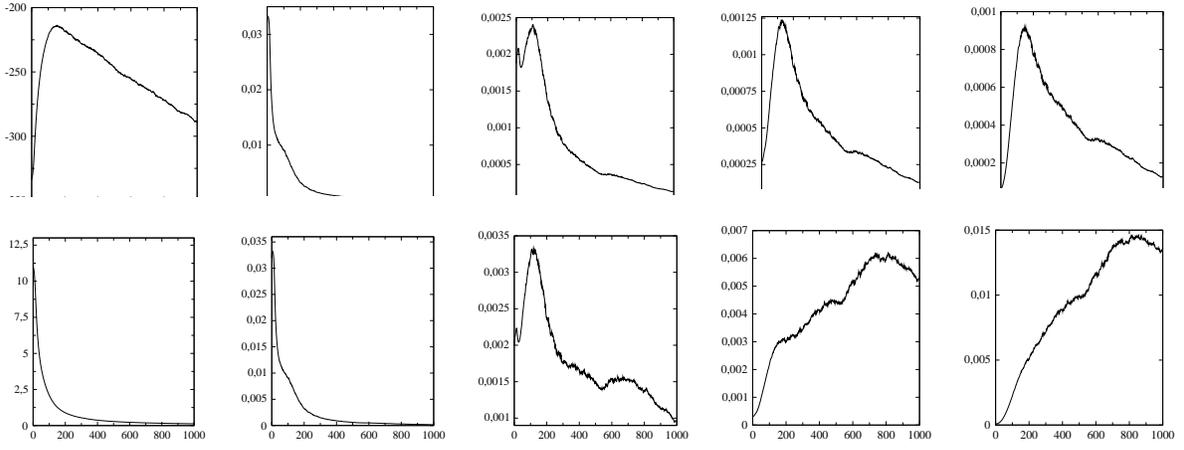

\begin{center}
\begin{tabular}{ccccc}
\includegraphics[width=0.15\textwidth] {BS_loglikelihood_Fig3_11.eps} &
\includegraphics[width=0.15\textwidth] {BS_xi_DA_d0_Fig3_12.eps} &
\includegraphics[width=0.15\textwidth] {BS_xi_DA_d1_Fig3_13.eps} &
\includegraphics[width=0.155\textwidth] {BS_xi_DA_d2_Fig3_14.eps} &
\includegraphics[width=0.155\textwidth] {BS_xi_DA_d3_Fig3_15.eps} \\
\includegraphics[width=0.145\textwidth] {BS_ProbRecErr_Fig3_21.eps} &
\includegraphics[width=0.15\textwidth] {BS_xi_DS_d0_Fig3_22.eps} &
\includegraphics[width=0.155\textwidth] {BS_xi_DS_d1_Fig3_23.eps} &
\includegraphics[width=0.155\textwidth] {BS_xi_DS_d2_Fig3_24.eps} &
\includegraphics[width=0.155\textwidth] {BS_xi_DS_d3_Fig3_25.eps}
\end{tabular}
\end{center}
\caption{Same as in figure~\ref{fig_RAN14} for the BS data set.}
\label{fig_BS_4x4}
\end{figure*}

\begin{figure*}[t!]
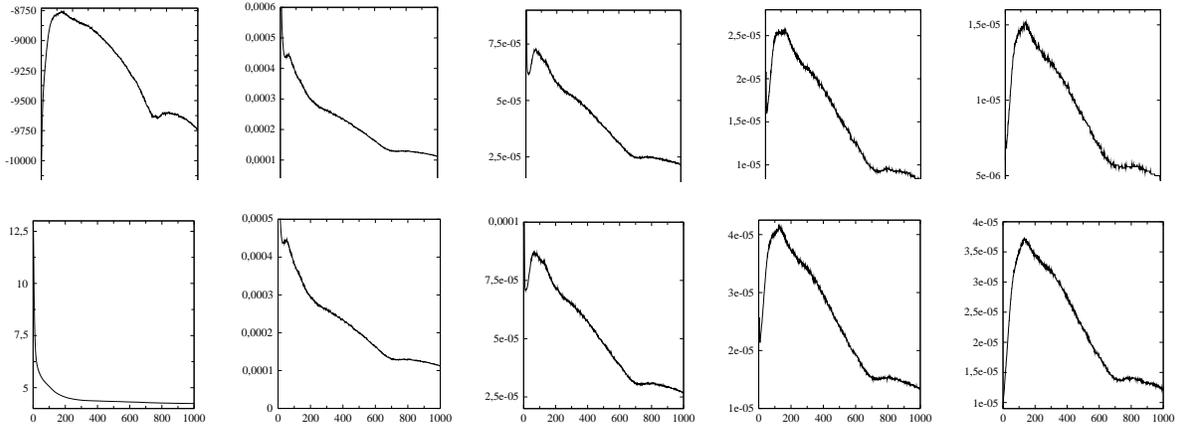

\begin{center}
\begin{tabular}{ccccc}
\includegraphics[width=0.15\textwidth] {LSE_loglikelihood_Fig4_11.eps} &
\includegraphics[width=0.15\textwidth] {LSE_xi_DA_d0_Fig4_12.eps} &
\includegraphics[width=0.15\textwidth] {LSE_xi_DA_d1_Fig4_13.eps} &
\includegraphics[width=0.15\textwidth] {LSE_xi_DA_d2_Fig4_14.eps} &
\includegraphics[width=0.15\textwidth] {LSE_xi_DA_d3_Fig4_15.eps} \\
\includegraphics[width=0.145\textwidth] {LSE_ProbRecErr_Fig4_21.eps} &
\includegraphics[width=0.155\textwidth] {LSE_xi_DS_d0_Fig4_22.eps} &
\includegraphics[width=0.155\textwidth] {LSE_xi_DS_d1_Fig4_23.eps} &
\includegraphics[width=0.15\textwidth] {LSE_xi_DS_d2_Fig4_24.eps} &
\includegraphics[width=0.155\textwidth] {LSE_xi_DS_d3_Fig4_25.eps}
\end{tabular}
\end{center}
\caption{Same as in figure~\ref{fig_RAN14} for the LSE data set.}
\label{fig_LSE}
\end{figure*}

Figure~\ref{fig_RAN10} shows our results for the RAN10 data set.  The
upper left panel shows the log-likelihood of the data during
training. As it can be seen, there is a clear maximum that should be
identified as the stopping point. The panels below show the
reconstruction errors~(\ref{reconstruction-error-RBM-probability})
and~(\ref{rec-error-squares}) which clearly fail to identify the
desired extremum. The rest of the columns show results for distances
$d=0,1,2$ and $3$. The first row depicts $\xi_d$ for $D_A$, where all
states at the required distances are taken into account.  As it can be
seen, starting at $d=1$ the criterion is robust and consistently
detects the maximum of the log-likelihood at the right place, thus
reinforcing the idea that the neighborhood of the data contains
valuable information.  The second row shows the denominator of $\xi_d$
corresponding to the first row, that is, the sum of probabilities of
the states included in each case. Notice that for $d=3$ this sum
equals one and $\xi_d$ is exactly equal to the likelihood of the
data. More interestingly, even when the sum is still far away from
one, as it happens for $d=1$, $\xi_d$ consistently finds the desired
point. This behavior is also observed in the rest of the data sets
analyzed.  Finally the third row shows $\xi_d$ for $D_S$, thus showing
the behavior of the criterion applied to different shells.  For $d=1$
and $2$ the criterion detects reasonably well the maximum of the
log-likelihood and can be used to identify the desired stopping
point. Notice, though, that the data alone, entirely contained at
$d=0$, is not capable to reproduce this behavior. Moreover, for $d$
larger than $2$ the criterion also fails, as it is expected that
starting at a certain distance the information regarding the model is
lost. Please notice that the initial transitory behavior of some of
the plots above is meaningless and can be omitted so it has been cut.

Equivalent results for the RAN14 case are shown in
figure~\ref{fig_RAN14}. The log-likelihood and the probabilistic
reconstruction error in~(\ref{reconstruction-error-RBM-probability})
are depicted in the upper and lower panels in the first column,
respectively. The other panels show $\xi_d$ for $D_A$ and $D_S$, with
$d=0,1,2,3$ (top and bottom rows, second to fifth columns). As in the
previous case, the reconstruction error fails to detect the maximum of
the likelihood, thus not being very useful in the present context.  On
the contrary, a stopping point obtained from $\xi_d$ selects a
near-optimal model. 
Notice that
the criterion is robust along all distances explored, as desired.
Similar results are found for the RAN12 and RAN16 cases. As it can be
inferred from these results, the optimal value of $d$ can not be fixed
beforehand and is problem-dependent.

The same plots for the BS and LSE problems are found in
figures~\ref{fig_BS_4x4} and~\ref{fig_LSE}. Once again, the
reconstruction error decreases monotonously and is therefore useless
in the present context, while $\xi_d$ for $d$ larger than 1
successfully does the task for $D=D_A$, while for $D=D_S$ the
criterion does not work in the BS problem.

\begin{table*}[t!]\centering
\begin{tabular}{|l|r|r|r|r|r|r|r|r|r|r|} \hline
\multicolumn{1}{|c}{Data Set}  & \multicolumn{10}{|c|}{Hamming Distance} \\ \hline
                               & 1 & 2 & 3 & 4 & 5 & 6 & 7 & 8 & 9 & 10 \\ \hline
{\em Bars and Stripes}         & 480 & 3216 & 11360 & 20744 & 19296 & 8688 & 1632 & 90 & - & - \\ \hline
{\em Labeled Shifter Ensemble} & 8434 & 41160 & 110326 & 165088 & 132976 & 54160 & 10368 & 966 & 40 & 2 \\ \hline
\end{tabular}
\caption{Number of neighbors at different Hamming distances for the
  BS and LSE data sets.}
\label{number-of-neighbours}
\end{table*}

\begin{figure*}[t!]
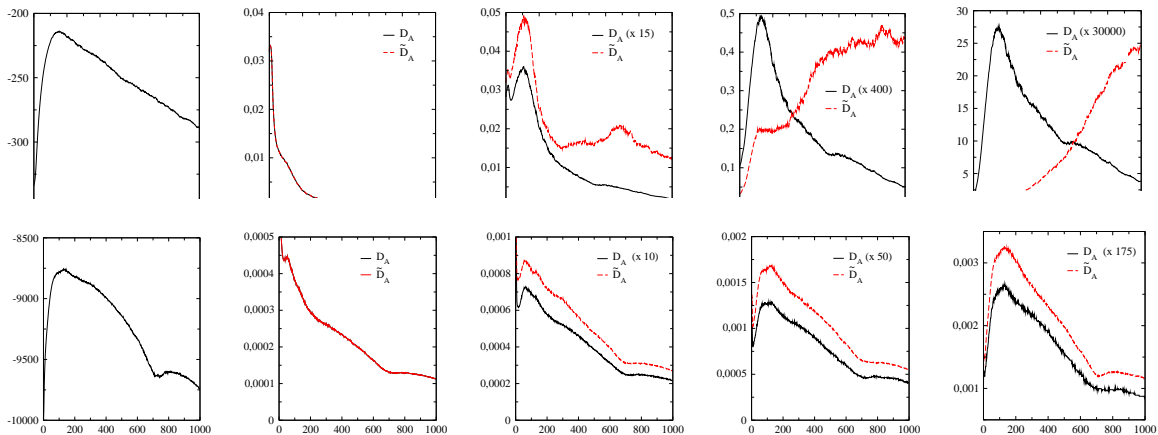

\begin{center}
\begin{tabular}{ccccc}
\includegraphics[width=0.15\textwidth]{BS_loglikelihood_FigLast_11.eps} &
\includegraphics[width=0.15\textwidth]{xi_tildexi_d0_BS_FigLast_12.eps} &
\includegraphics[width=0.15\textwidth]{xi_tildexi_d1_BS_FigLast_13.eps} &
\includegraphics[width=0.145\textwidth]{xi_tildexi_d2_BS_FigLast_14.eps} &
\includegraphics[width=0.145\textwidth]{xi_tildexi_d3_BS_FigLast_15.eps} \\
\includegraphics[width=0.15\textwidth]{LSE_loglikelihood_FigLast_21.eps} &
\includegraphics[width=0.15\textwidth]{xi_tildexi_d0_LSE_FigLast_22.eps} &
\includegraphics[width=0.15\textwidth]{xi_tildexi_d1_LSE_FigLast_23.eps} &
\includegraphics[width=0.15\textwidth]{xi_tildexi_d2_LSE_FigLast_24.eps} &
\includegraphics[width=0.15\textwidth]{xi_tildexi_d3_LSE_FigLast_25.eps}
\end{tabular}
\end{center}
\caption{Comparison between $\xi_d$ (black curves) and $\tilde\xi_d$
  (red curves) for the BS and LSE data sets (upper and lower
  rows). Notice that since the magnitude of these parameters is
  irrelevant, some curves have been scaled for the sake of clarity.
  The first column plots the log-likelihood of the data along the
  simulation.}
\label{fig_BS_LSE_sales}
\end{figure*}

\begin{figure*}[t!]
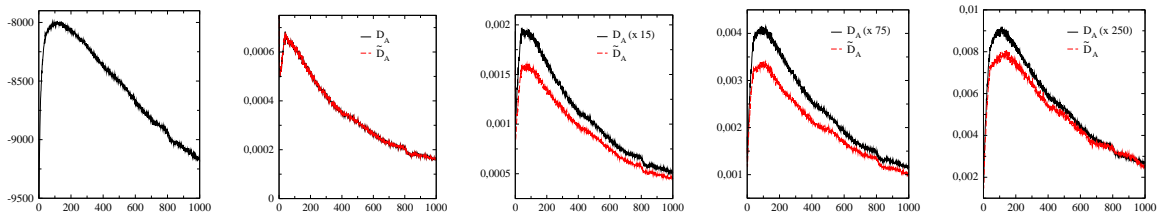

\begin{center}
\begin{tabular}{ccccc}
\includegraphics[width=0.15\textwidth]{LSE_loglikelihood_FigCD10_11.eps} &
\includegraphics[width=0.15\textwidth]{LSE_xi_tildexi_d0_FigCD10_12.eps} &
\includegraphics[width=0.15\textwidth]{LSE_xi_tildexi_d1_FigCD10_13.eps} &
\includegraphics[width=0.15\textwidth]{LSE_xi_tildexi_d2_FigCD10_14.eps} &
\includegraphics[width=0.15\textwidth]{LSE_xi_tildexi_d3_FigCD10_15.eps}
\end{tabular}
\end{center}
\caption{Same as in figure~\ref{fig_BS_LSE_sales} for the LSE problem in CD$_{10}$.}
\label{fig_LSE_CD10}
\end{figure*}

\subsection{Uncomplete Neighborhoods}
\label{uncomplete}

Despite the success of the criterion built as proposed, it is clear
that for large spaces it can be unpractical if the number of states in
the neighborhood of the training set is very large. For that reason,
we have tested the criterion on randomly selected subsets $\tilde D_A
\subset D_A$ of the same size as the training set, which is always
computationally tractable. In this sense, we denote by $\tilde \xi_d$
the evaluation of $\xi_d$ on $\tilde D_A$.
Figure~\ref{fig_BS_LSE_sales} shows $\tilde \xi_d$ compared with
$\xi_d$ from the previous figures for the BS (first row) and LSE
(second row) problems. More precisely, the first column shows the
log-likelihood of the data along the training process, while the rest
of the columns plot both $\tilde\xi_d$ and $\xi_d$ for $d=0,1,2$ and
$3$. Notice that the absolute scales of $\xi_d$ and $\tilde\xi_d$ may
vary mainly due to the value of the sum of probabilities in the
denominators. However, since the precise value of these quantities is
irrelevant, we have decided to scale them properly for the sake of
comparison. Although $\tilde\xi_d$ is built from a much smaller set
than $\xi_d$, it captures all the significant features of $\xi_d$ and
can therefore be used instead of it. In this sense, $\tilde\xi_d$
provides a good stopping criterion for CD$_1$, although it is not as
robust as $\xi_d$ due to the strong reduction of states contributing
to $\tilde\xi_d$ as compared with those entering in $\xi_d$.  This
reduction is illustrated in table~\ref{number-of-neighbours}, where we
show the number of neighboring states to the data set at different
distances for the BS and LSE problems.  By increasing the number of
states included in $\tilde\xi_d$, convergence to $\xi_d$ is expected
at the expense of an increase in computational cost. However, the
present results indicate that, at least for the problems at hand, a
number of examples similar to that of the training set in the
evaluation of $\tilde\xi_d$ is enough to detect the maximum of the
log-likelihood of the data.

All the results presented up to this point show the goodness of the
proposed stopping criterion for learning in CD$_1$. However, the
underlying idea can be applied to different learning algorithms that
try to maximize the log-likelihood of the data. In this way we have
repeated all the previous experiments for CD$_{10}$ with very similar
results to the ones above. As an example, figure~\ref{fig_LSE_CD10}
shows the log-likelihood, $\xi_d$ and $\tilde \xi_d$ with $d=0,1,2,3$
and CD$_{10}$ for the LSE data set, which is the largest one analyzed
in this work. As it is clearly seen, the quality of the results is
very similar to the CD$_1$ case, thus stressing the robustness of the
criterion.

\begin{figure*}[t!]\centering
\includegraphics[width=1.0\textwidth] {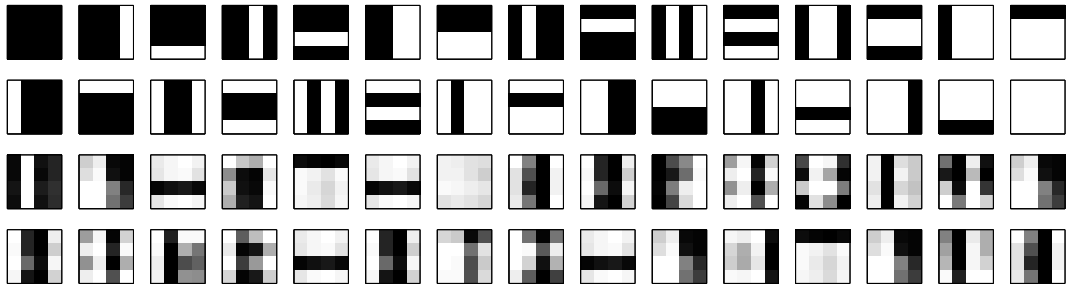}
\caption{ Training data (two upper rows) and generated samples (two
  lower rows) for the BS problems with the weights and bias obtained
  at the stopping point detected by $\tilde\xi_d$ with $d=1$.}
\label{barritas}
\end{figure*}

As a final remark, we note that for the BS problem the trained RBM
stopped using the proposed criterion is able to qualitatively generate
samples similar to those in the training set. We show in
figure~\ref{barritas} the complete training set (two upper rows) and
the same number of generated samples (two lower rows) obtained from
the RBM trained with CD$_1$ and stopped after 5000 epochs, around the
maximum shown by $\tilde\xi_{d=1}$, which approximately coincides with
the optimal value of the log-likelihood. It is important to realize
that, ultimately, the quality of the model is a direct measure of the
quality of CD$_1$ learning, and that the model used to generate the
plots is the one with largest $\tilde\xi_d$, which is quite close to
the one with largest likelihood.

\section{Conclusions}

In this work we have introduced the contribution of neighboring
points to the training set to build a stopping criterion for learning
in CD$_1$. We have shown that not only the training set but also the
neighboring states contain valuable information that can be used to
follow the evolution of the network along training.

Based on the fact that learning tries to increase the contribution of
the relevant states while decreasing the contribution of the rest,
continuity and smoothness of the energy function assigns more
probability to states close to the training data.  This is the key
idea behind the proposed stopping criterion.  In fact, two different
but related estimators (depending on the number of states used to
compute them) have been proposed and tested experimentally. The first
one includes all states close to the training set, while the second
one takes only a fraction of these states as small as the size of the
training set.  The first estimator is robust but may require from the
use of a forbiddingly large amount of states, while the second one is
always tractable and captures most of the features of the first one,
thus providing a suitable stopping learning criterion.  This second
estimator could be used in larger data set problems, where an exact
computation of the log-likelihood is not possible. Additionally, the
main idea of proximity to the training set will be explored in other
aspects related to learning in future work.

\section*{Acknowledgments} 

ER: This research is partially funded by Spanish research project
TIN2012-31377.
 
FM: This work has been supported by grant No. FIS2014-56257-C2-1-P
from DGI (Spain).

JD: This work was partially supported by SGR2014-890 (MACDA) of the
Generalitat de Catalunya, MICINN project BASMATI
(TIN2011-27479-C04-03) and MINECO project APCOM (TIN2014-57226-P)

\bibliographystyle{IEEEtran}

\end{document}